\documentclass[11pt]{article}  

\usepackage[margin=1.5in]{geometry}
\usepackage{graphicx}
\usepackage{epstopdf}
\usepackage{amsfonts}
\usepackage{amssymb}
\usepackage{subfigure}
\usepackage{amsmath,bm}
\usepackage{tikz}
\usepackage{color}
\usepackage{adjustbox}
\usetikzlibrary{shapes,arrows,chains}
\usetikzlibrary[calc]
\usetikzlibrary{positioning}
\usepackage{algorithm}
\usepackage{cite}
\usepackage{verbatim}
\usepackage[noend]{algpseudocode}
\newcounter{casenum}

\renewcommand{\Re}{\mathbb{R}}
\newcommand{\e}{\,\mathrm{e}}

\title{\LARGE \bf
Hierarchical State Abstractions for Decision-Making Problems\\ with Computational Constraints
}

\date{~}

\author{Daniel T. Larsson\thanks{D. Larsson is a PhD student with the D. Guggenheim School of Aerospace Engineering, Georgia Institute of Technology, Atlanta,
		GA, 30332-0150, USA. Email:
		{\small daniel.larsson@gatech.edu}}
	, Daniel Braun\thanks{D. Braun is a Professor with the Institute of Neural Information Processing,
		Ulm University, D-89081 Ulm, Germany. Email:
		{\small daniel.braun@uni-ulm.de}}
	, Panagiotis Tsiotras\thanks{P. Tsiotras is a Professor with the D. Guggenheim School of Aerospace Engineering and the Institute for Robotics and Intelligent Machines, Georgia Institute of Technology, Atlanta,
		GA, 30332-0150, USA. Email:
		{\small tsiotras@gatech.edu}
		\newline
	\indent The work of DL and PT has been supported by ONR award N00014-13-1-0563 and of DB by project ERC STG BRISC 678082.}
}

\begin{document}

\renewcommand{\baselinestretch}{1.0}

\maketitle
\thispagestyle{empty}
\pagestyle{empty}


\begin{abstract}
	In this semi-tutorial paper, we first review the information-theoretic approach to account
for the computational costs incurred during the search for optimal actions in a sequential decision-making problem.
	The traditional (MDP) framework ignores computational limitations while searching for optimal policies, essentially
assuming that the acting agent is perfectly rational and aims for exact optimality.
Using the  free-energy, a variational principle is introduced that accounts not only for the value of a policy alone, but also considers the cost of finding this optimal policy.
The solution of the variational equations arising from this formulation can be obtained using familiar Bellman-like value iterations from dynamic programming (DP) and the Blahut-Arimoto (BA) algorithm from rate distortion theory.
Finally,
we demonstrate the utility of the approach for generating hierarchies of state abstractions that can be used to best exploit the available computational resources.
A numerical example showcases these concepts for a path-planning problem in a grid world environment.
\end{abstract}

\section{INTRODUCTION}
\label{sec:intro}

Markov Decision Processes (MDP) is the standard framework for optimal sequential decision making under uncertainty~\cite{SuttonBarto}.
However, this framework assumes that the decision maker, customary referred to as the ``agent,'' is perfectly rational \cite{Lipman95}.
That is, the agent is assumed to have unlimited computational and/or time resources to process all information and compute the optimal solution unhindered.
In many practical applications  this is a strong and often unrealistic assumption.
For example, biological systems are {adaptive} in nature and excel at finding solutions to complex problems that may not be globally optimal but  are ``good enough" given the constraints and/or available sensory information~\cite{Genewein15,Lipman95}.

Over the past several years considerable effort has been devoted to developing a framework
that accounts for the available computational constraints.
The idea is to account for the cost of computation
not after the fact, as is traditionally the case, but from the beginning, as part of the problem formulation.
This steam of research has
led to the development of \textit{information-constrained MDPs}~\cite{Tishby11,GrauMoya16,Genewein15} that utilize concepts from information theory in order to
model the computational limitation(s) of the agent~\cite{Ortega11,Rubin12}.
The approach also has strong connections with the thermodynamic point of view of the theory of computation~\cite{bennett1982thermodynamics,feynman1998feynman,Ortega13}.

The introduction of the added constraints (computational and otherwise) to the decision process is meant to model an agent that does not have access to unlimited resources, and hence will select actions that are not optimal in the traditional sense of maximizing value or utility, but rather in the sense of maximizing (expected)
value or utility \emph{for a given processing cost}~\cite{Lipman95,Tishby11,Rubin12}.
Such a decision maker, who alters its behavior based on its resources, is said to be a \emph{bounded rational}
agent \cite{Lipman95,Simon1982,Evang11,horvitz2001computational}.
In this work, we apply this framework to develop decision-making algorithms for path-planning problems while also accounting for the information
processing costs of computing the optimal {actions}.

The information-limited MDP problem is formally posed as a constrained optimization problem, where the objective is to maximize the value function
subject to a constraint that penalizes the difference between the resulting policy and a known prior (computationally cheap) policy.
This constraint is often represented using the Kullback-Leibler (KL) divergence~\cite{Tishby11,Rubin12,Genewein15}.
The unconstrained optimization problem obtained by introducing Lagrange multipliers likens that of the free-energy function from thermodynamics~\cite{Ortega13} and is hence called the ``free energy" of the information limited decision process~\cite{Tishby11,Rubin12,Ortega13,Evang11}.
In this formulation the goal is to find a policy that maximizes the free energy at each step of the process.
The solution to this problem can be found numerically through value-iteration type of algorithms or backward recursion \cite{BertsekasV207,Rubin12,Tishby11} and follows closely the process for finding the optimal value function using dynamic programming (DP) \cite{SuttonBarto,BertsekasV207,Bellman57}.

The framework can be augmented so that, across \emph{all states on average}, the transformation (information) cost of ``molding" this prior into the optimal policy is reduced~\cite{Tishby11,Genewein15}.
The resulting optimization problem is analogous to that of the Blahut-Arimoto algorithm used extensively in rate distortion theory~\cite{Tishby11,Tishby99}.

The concept of mutual information between states and actions is important and is a topic that does not seem to have been exploited extensively in the controls community.
Most importantly, it leads to a direct analogy of the amount of information an action carries about the state.
For example, a perfectly rational decision maker will solve the optimization problem producing a policy that is \emph{state specific}.
This means knowing what action was selected is very informative regarding what state the agent was in when action was taken~\cite{Genewein15}.
However, a bounded rational agent must consider the resources required to find the optimal action and hence finds policies that are relatively generic and not state-specific.
This classification of more informative versus less informative states leads in a natural way to a series of state space \emph{abstractions}, which are discussed in greater detail in this paper.
Such abstractions can be utilized in order to focus the available computational resources in a top-down, task-specific manner.

The use of hierarchical abstractions has been explored to find efficient solutions to path-planning problems~\cite{CowTsi:acc08,CowTsi:cdc10b,CowTsi:smc11,TJB:jirs11,CowTsi:tro11,Behnke04,KaDa86} subject to computational constraints.
In those references, hierarchies of multi-resolution abstractions were created to speed-up replanning.
The difference with the current approach is that while in those references these abstractions were imposed externally (and perhaps rather axiomatically) by the designer, here they arise intrinsically, as the result of the limited available computational resources of the agent.

The scope of this paper is twofold.
First, we provide an overview of the existing framework of the variational free-energy principle that allows the construction of state abstractions, which can then be utilized to solve sequential decision making problems subject to computational constraints.
Second, we propose a novel modification to the Blahut-Arimoto algorithm, allowing us to handle a closed-loop optimization problem with the existing framework.
Finally, the proposed approach is demonstrated on a typical path-planning problem in a grid world.
This is the first work, as far as the authors know, where the information-theoretic bounded rationality framework is used for multi-resolution hierarchical abstraction in sequential decision problems.


\section{PROBLEM FORMULATION}        \label{sec:prob_formul}

\subsection{Preliminaries of Standard Markov Decision Process}

We consider sequential decision-making problems in the presence of uncertainty.
Markov Decision Processes (MDPs) is the standard mathematical framework to solve such problems in discrete time.
We will use $x_t$ to denote the system state and $u_t$ the control input at time $t$, where $t=0,1,2,\ldots$.
The system state is an element of the state set $X$ ($x_t \in X$) and the control input is selected from the set of admissible inputs, $u_t \in U(x_t)$.
The system dynamics are modeled via a collection of
transition probabilities  $P(x_{t+1}|x_{t},u_{t})$, which represent the probability that the agent, being at state $x_t$ and selecting input $u_t$ at time step $t$, will find itself at state $x_{t+1}$ at the next instance of time.
Each action results in a reward $R(x_{t+1},x_{t},u_{t})$ which the agent receives for selecting input $u_t$ while at state $x_t$ and at time $t$ and transitioning to state $x_{t+1}$ at time $t+1$.

The objective of the infinite-time MDP is to find a policy which maximizes the future discounted expected reward.
By policy we mean a map that for each state $x_t$ provides a probabilistic distribution over actions, $\pi(u_t|x_t)$.
Since this paper is limited to infinite horizon problems, we assume that our policy is stationary, that is, $\pi$ is independent of $t$.

More formally, we seek to maximize  the objective function
\begin{equation}    \label{eq:policyvalue1}
V^{\pi}(x_0) \triangleq \lim_{T \to \infty}\mathbb{E}_\tau\Bigg[\sum^{T-1}_{t=0}\gamma^{t}R(x_{t+1},x_{t},u_t)  \Bigg],
\end{equation}
where $\gamma \in [0,1)$ is the \emph{discount factor}, $x_0$ is the initial state and $u_t$ is selected according to the provided policy $\pi$ at each time step.
Equation \eqref{eq:policyvalue1} can be interpreted as the value-to-go provided the system starts at state $x_0$ and executes the stationary policy $\pi$.
The expectation in equation \eqref{eq:policyvalue1} is over all future trajectories  $\tau = (x_0,u_0,x_1,u_1,\ldots,x_T)$ of length $T$ starting at $x_0$, which under the Markov property assumption is given as
\begin{equation}   \label{eq:trajectory1}
\mathrm{Pr}(u_{0},x_1,u_1,x_2,\ldots,x_T|x_0) = \prod_{t=0}^{T}\pi(u_t|x_t)P(x_{t+1}|x_t,u_t).
\end{equation}
The objective is then to find the optimal distribution over admissible inputs at each state, in order to maximize the objective function in \eqref{eq:policyvalue1}.
Introducing the shorthand notation $x = x_t$, $u = u_t$, and $x' = x_{t+1}$,
expanding \eqref{eq:policyvalue1} and making use of \eqref{eq:trajectory1}, we obtain
\begin{equation}    \label{eq:policyvalue3}
V^{\pi}(x) =
\sum_{u \in U(x)}\pi(u|x)\sum_{x' \in X}P(x'|x,u)\Big[
R(x',x,u) +\gamma{}V^{\pi}(x') \Big].
\end{equation}
We then write the optimal value function as~\cite{Rubin12}
\begin{equation}
V^{*}(x) = \max_{\pi}V^{\pi}(x),
\end{equation}
which can be equivalently written as
\begin{equation}
V^{*}(x) = \max_{u \in U(x)}\sum_{x' \in X}P(x'|x,u)\Big[R(x',x,u) + \gamma{}V^{*}(x') \Big],
\label{eq:opt_valuefunc}
\end{equation}
with the optimal policy given by $\pi^* = \mathrm{argmax}_{\pi}V^{\pi}(x)$.

We also define the {state-action} value function for {any policy} $\pi$ as \cite{SuttonBarto}
\begin{equation}
Q^{\pi}(x,u) \triangleq \sum_{x' \in X}P(x'|x,u)\Big[R(x',x,u) + \gamma{} V^{\pi}(x')\Big].
\label{eq:stateactionvalue3}
\end{equation}
Similarly, the {optimal} state-action value function is defined as
\begin{equation}
Q^{*}(x,u) \triangleq \sum_{x' \in X}P(x'|x,u)\Big[R(x',x,u) + \gamma{} V^{*}(x')\Big].
\label{eq:optstateactionvalue1}
\end{equation}
In the case of a {perfectly rational} decision maker, the agent will {deterministically} select the optimal action that maximizes equation \eqref{eq:optstateactionvalue1}.
That is, the agent will act greedy with respect to $Q^{*}(x,u)$.    
In this paper, we distinguish between a general {stochastic} policy and one that is {deterministic} by denoting optimal deterministic policies as $\Gamma^{*}(u|x)$ and all other optimal policies as $\pi^{*}(u|x)$.
Then, a perfectly rational agent will choose the policy
$
\Gamma^{*}(u|x) = \delta(u - u^*(x)),
$
where   $u^*(x) = \text{arg}\max_{u}Q^{*}(x,u)$.

\subsection{The Information-Limited Markov Decision Process}      \label{sec:IIb}

The problem formulation presented above does not consider the ``effort" required to find the optimal policy $\Gamma^{*}(u|x)$,
which maximizes \eqref{eq:policyvalue1}~\cite{Rubin12,Tishby11}.
In order to model the effort required to obtain the (global) maximum, we provide the agent with a prior choice distribution ($\rho(u|x)$) and then limit the amount by which the posterior policy ($\pi(u|x)$) is permitted to differ from $\rho(u|x)$ \cite{Genewein15,Tishby11,Rubin12}.
The difference between the two distributions is measured using the Kullback-Leibler ($\mathrm{KL}$) divergence~\cite{InformationThe06}
\begin{equation}
D_{\mathrm{KL}}\big(\pi(u|x)\|\rho(u|x)\big) \triangleq \sum_{u \in U(x)}\pi(u|x)\log\frac{\pi(u|x)}{\rho(u|x)}.
\end{equation}
The total discounted {information} cost for a decision maker starting from any initial state $x$ and following the policy $\pi(u|x)$ is then defined as \cite{Rubin12}
\begin{equation}   \label{eq:infodef}
\begin{aligned}
D^{\pi}(x)
& \triangleq \lim_{T \to \infty}\mathbb{E}_
\tau\Bigg[\sum_{t=0}^{T-1}\gamma^{t}\log\frac{\pi(u_t|x_t)}{\rho(u_t|x_t)} \Bigg],
\end{aligned}
\end{equation}
where the expectation in equation \eqref{eq:infodef} is with respect to the resulting trajectory, given by equation \eqref{eq:trajectory1}.
The goal is now to not only maximize the value alone, but maximize the trade-off between the value and information cost \cite{Rubin12,Tishby11,Genewein15,Evang11}.
This is a constrained optimization problem, which can be solved using Lagrange multipliers.
The free energy for a given policy $\pi(u|x)$ is defined as
\begin{equation}      \label{eq:FreeEnergy}
F^{\pi}(x;\beta) \triangleq V^{\pi}(x) - \frac{1}{\beta}D^{\pi}(x),
\end{equation}
where $\beta > 0$.
The objective is now to find the policy $\pi^{*}(u|x)$ which maximizes equation \eqref{eq:FreeEnergy}.
Substituting known quantities and expanding terms, we obtain the mapping $\mathcal{B}: \Re^{|X|} \to \Re^{|X|}$ defined as
\begin{equation}   \label{eq:FEMap}
\begin{aligned}
&\mathcal{B}[F](x;\beta) \triangleq \max_{\pi(u|x)}\sum_{u \in U(x)}\pi(u|x)\times \\
&\Big[\sum_{x' \in X}P(x'|x,u)R(x',x,u) - \frac{1}{\beta}\log{\frac{\pi(u|x)}{\rho(u|x)}}  \\
 &~~~~~~~~~~~~~~ + \gamma{}\sum_{x' \in X}P(x'|x,u)F(x';\beta) \Big].
\end{aligned}
\end{equation}
The proof of how this mapping results from equation \eqref{eq:FreeEnergy} can be found in \cite{Rubin12}.
The optimal value of the free energy $F^*(x;\beta)$ is the fixed point of the equation $F^* = \mathcal{B}[F^*]$.
Similarly to (\ref{eq:stateactionvalue3}), we define
\begin{equation}    \label{eq:QF}
Q_{F}(x,u;\beta) \triangleq \sum_{x' \in X}P(x'|x,u)\Big[R(x',x,u) + \gamma{}F(x';\beta)\Big],
\end{equation}
and
$Z(x;\beta) \triangleq \sum_{u \in U(x)}\rho(u|x)\e^{\beta{}Q_{F}(x,u;\beta)}$.
As in \cite{Rubin12,Ortega13}, equation \eqref{eq:FEMap} can be shown to be equivalent to
\begin{equation} \label{eq:FEMap_eq}
\mathcal{B}[F](x;\beta) = \frac{1}{\beta}\log{Z(x;\beta)}.
\end{equation}
This is iteratively applied until convergence to obtain the optimal free energy $F^{*}(x;\beta)$.
For a given value of $\beta$ the optimal policy for the infinite horizon problem is then
\begin{equation}
\pi^{*}(u|x) = \frac{\rho(u|x) \e^{\beta{Q^{*}_F(x,u;\beta)}}}{Z^*(x;\beta)},
\label{eq:optimalpolicyFE}
\end{equation}
where $Q^{*}_{F}(x,u;\beta)$ and $Z^{*}(x;\beta)$ are given by
\begin{equation}
\begin{aligned}
Q^{*}_{F}(x,u;\beta) &= \sum_{x' \in X}P(x'|x,u)\Big[R(x',x,u) + \gamma{}F^{*}(x';\beta)\Big], \\
Z^{*}(x;\beta) &= \sum_{u \in U(x)}\rho(u|x)\e^{\beta{}Q^{*}_{F}(x,u;\beta)}.
\end{aligned}
\end{equation}
Using this framework we are able to model a wide variety of agents with limited resources by varying the resource parameter, $\beta$.
Hence, $\beta$ can be viewed as a parameter which reflects the agents computational abilities \cite{GrauMoya16}.
We now discuss the limiting cases of the resource parameter, $\beta$.

	\textbf{Case~I} ($\beta \to 0$):{~~In this case the information term in equations \eqref{eq:FreeEnergy} and \eqref{eq:FEMap} becomes dominant and the process becomes mainly concerned with reducing the information cost $D^{\pi}(x)$.
	Note that the $\mathrm{KL}$ divergence is always nonnegative-definite \cite{InformationThe06} and  hence obtains its minimum at zero.
	This leads to $\pi^{*} = \rho$.
	That is, the prior distribution is directly ``copied" to the posterior policy and represents an agent that is not able to deviate from its initial notion of an optimal action, indicative of a decision maker with no computational resources \cite{GrauMoya16}.}
	
	\textbf{Case~II} ($\beta \to \infty$):{~~In the limit as $\beta \to \infty$ the information cost in equations \eqref{eq:FreeEnergy} and \eqref{eq:FEMap} is effectively removed.
	The process will focus on finding a policy that maximizes $V^{\pi}(x)$.
	This means that the agent will behave more like its non-information limited counterpart.
	We therefore expect to recover the non-information limited policy in the limit (i.e., $\pi^{*} \to \Gamma^{*}$), which is indicative of an agent that has unlimited computational resources \cite{GrauMoya16}.}

\section{The Information Limited MDP and the Modified Blahut-Arimoto Algorithm}  \label{secIII}

In this section we address the question of whether we can find an optimal a priori action distribution $\rho(u|x)$ so as to minimize the information cost across all states \emph{on average}.
Note that the information limited MDP does not modify $\rho(u|x)$ when searching for the optimal policy $\pi^{*}$.
Instead, it will find the policy that best trades the value and information for a given $\rho(u|x)$.
The problem now becomes one of finding both the posterior policy and prior action distribution so as to maximize equation \eqref{eq:FreeEnergy} on average across all system states.

Averaging equation \eqref{eq:FreeEnergy} over system states and assuming a state-independent prior distribution over actions ($\rho(u|x) = \rho(u)$), we obtain a relation analogous with that of the rate distortion function from information theory \cite{Tishby99,Genewein15,Rubin12,Tishby11}.
The relation between equation \eqref{eq:FEmapOpenloop2} and that of the rate distortion function in \cite{Tishby99} is quite intimate, as the averaging over system states of \eqref{eq:FreeEnergy} gives rise to the direct appearance of the mutual information between actions and states \cite{Genewein15,InformationThe06}.
Optimizing the averaged free energy over both $\pi$ and $\rho$ we arrive at the Blahut-Arimoto (BA) algorithm.
Formally, the problem can be stated as
\begin{equation}
\max_{\rho}\sum_{x \in X}p(x)
\Big(\max_{\pi}\sum_{u \in U(x)}\pi(u|x)\Big[Q_{F}(x,u;\beta) - \frac{1}{\beta}\log{\frac{\pi(u|x)}{\rho(u)}}\Big]\Big).
\label{eq:FEmapOpenloop2}
\end{equation}
Optimizing equation \eqref{eq:FEmapOpenloop2} with respect to $\rho(u)$ is a convex optimization problem \cite{Genewein15,InformationThe06}, and is numerically solved by recursively iterating between the following relations until convergence \cite{Tishby11,Genewein15,Tishby99}
\begin{align}
\pi^{*}(u|x) &= \frac{\rho^*(u) \e^{\beta{Q^{*}_F(x,u;\beta)}}}{\sum_{u}\rho^*(u)\e^{\beta{}Q^{*}_{F}(x,u;\beta)}},  \label{eq:BAbasicsA} \\
\rho^*(u) &= \sum_{x}p(x)\pi^{*}(u|x).  \label{eq:BAbasicsB}
\end{align}
Here $p(x)$ is the (fixed) probability distribution over states and $\pi^{*}$ is found in the same manner as described in the previous section.
The proof can be found in \cite{Genewein15,Tishby11}.

Similarly to the previous section, an agent utilizing the BA algorithm for path planning will find a policy that deviates in various amounts to the provided prior action distribution depending on the value of $\beta{}$.
Now, however, the prior action distribution is \emph{not} static and is updated each iteration to be consistent with the posterior policy, $\pi$ (see equations \eqref{eq:BAbasicsA}-\eqref{eq:BAbasicsB}).
Note that at low $\beta{}$ the agent becomes mainly concerned with minimizing the mutual information between states and actions, as given by $\sum_{x\in X}p(x)D^{\pi}(x)$ in \eqref{eq:FEmapOpenloop2}.
This results in a policy that is generic with respect to the state, leading to {state abstractions} as shown in \cite{Genewein15}.
That is, the agent will focus its efforts to find a single action that yields a high reward on average across the state space.

This is no issue for agents that have access to all actions in every state or that have enough resources (high $\beta{}$) to sufficiently alter $\rho$ into $\pi$.
However, for agents that are resource limited and have admissible actions spaces that are state dependent, this behavior may lead to a posterior policy that assigns non-zero probability to actions that are not available in a certain states.
To remedy this issue, we propose an extension to the BA algorithm by altering the update equations in \eqref{eq:BAbasicsA}-\eqref{eq:BAbasicsB} as follows

\begin{equation}
\begin{aligned}
\pi^{*}(u|x) &= \frac{\rho^*(u|x) \e^{\beta{Q^{*}_F(x,u;\beta)}}}{\sum_u\rho^*(u|x) \e^{\beta{Q^{*}_F(x,u;\beta)}}}, \\
\hat{\pi}^*(u) &= \sum_{x}p(x)\pi^{*}(u|x), \\
\rho^*(u|x) &= \frac{\hat{\pi}^*(u) \e^{Q_{p}(x,u)}}{\sum_{u}\hat{\pi}^*(u) \e^{Q_{p}(x,u)}}.
\end{aligned}
\label{eq:BaBasics2}
\end{equation}
By introducing the penalty function $Q_{p}(x,u)$, we ensure that the agent may only deliberate among available actions in each state while forming its posterior policy $\pi^{*}$.
The penalty function can then be constructed so that $Q_{p}(x,u) \to -\infty$ for actions that are inadmissible in a given state.
In the numerical example that follows, we use $Q_{p}(x,u) = \pm100$ for actions that are available/unavailable, respectively.

\section{NUMERICAL EXAMPLE}     \label{sec:results}

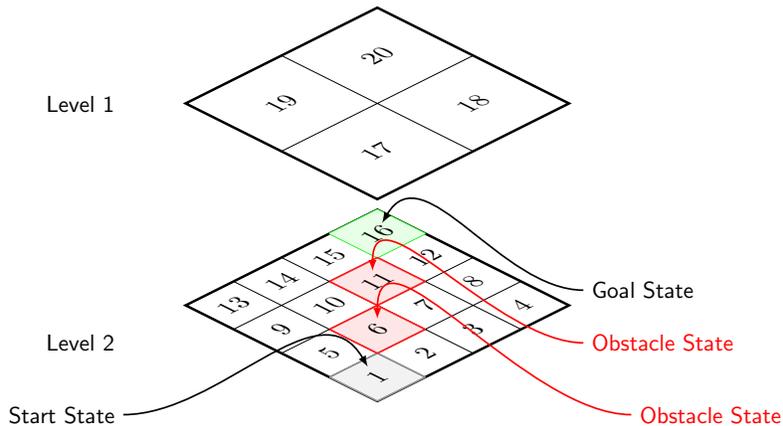
\begin{figure}[h!]
	\centering
	\begin{adjustbox}{max size={0.75\textwidth}}
		\begin{tikzpicture}[scale=.8,every node/.style={minimum size=1cm},on grid]
		
		\begin{scope}[
		yshift=-120,every node/.append style={
			yslant=0.5,xslant=-1},yslant=0.5,xslant=-1
		]
		\fill[white,fill opacity=0.9] (0,0) rectangle (4,4);
		\draw[step=1cm, black] (0,0) grid (4,4); 
		\draw[black,very thick] (0,0) rectangle (4,4);
		\draw[step=1cm, green!90,thick] (3,3) grid (4,4);
		\fill[green!10!] (3,3) rectangle (4,4);

		\fill[gray!10!] (0,0) rectangle (1,1);
		\draw[step=1cm, gray!90,thick] (0,0) grid (1,1);

		\fill[red!10!] (1,1) rectangle (2,2);
		\draw[step=1cm, red!90,thick] (1,1) grid (2,2);
		\fill[red!10!] (2,2) rectangle (3,3);
		\draw[step=1cm, red!90,thick] (2,2) grid (3,3);
		
		\node at (0.5,0.5) {1};
		\node at (1.5,0.5) {2};
		\node at (2.5,0.5) {3};
		\node at (3.5,0.5) {4};
		\node at (0.5,1.5) {5};
		\node at (1.5,1.5) {6};
		\node at (2.5,1.5) {7};
		\node at (3.5,1.5) {8};
		\node at (0.5,2.5) {9};
		\node at (1.5,2.5) {10};
		\node at (2.5,2.5) {11};
		\node at (3.5,2.5) {12};
		\node at (0.5,3.5) {13};
		\node at (1.5,3.5) {14};
		\node at (2.5,3.5) {15};
		\node at (3.5,3.5) {16};
		
		\end{scope}
		
		\begin{scope}[
		yshift=0,every node/.append style={
			yslant=0.5,xslant=-1},yslant=0.5,xslant=-1
		]
		\fill[white,fill opacity=.9] (0,0) rectangle (4,4);
		\draw[black,very thick] (0,0) rectangle (4,4);
		\draw[step=2cm, black] (0,0) grid (4,4);
		\node at (1,1) {17};
		\node at (3,1) {18};
		\node at (1,3) {19};
		\node at (3,3) {20};
		\end{scope}

		
		\node at (-6.2,2) {$\mathsf{Level~1}$};
		\node at (-6.2,-3) {$\mathsf{Level~2}$};

		\draw[-latex,thick,red](5.3,-4.5)node[right]{$\mathsf{Obstacle~State}$}
		to[out=180,in=90] (0,-2.5);
		
		\draw[-latex,thick,black](4.3,-1.9)node[right]{$\mathsf{Goal~State}$}
		to[out=180,in=60] (0.1,-0.5);
		
		\draw[-latex,thick,red](4.3,-3)node[right]{$\mathsf{Obstacle~State}$}
		to[out=180,in=90] (-0.1,-1.5);
		
		\draw[-latex,thick,black](-5.3,-4.5)node[left]{$\mathsf{Start~State}$}
		to[out=0,in=120] (-0.2,-3.5);
		\end{tikzpicture}
	\end{adjustbox}
	\caption{Grid layering with state numbering.  Start state indicated in gray, obstacle states in red with goal state in green.}
	\label{fig:gridex11}
\end{figure}

\begin{figure}[htb]
	\centering
	\begin{adjustbox}{max size={0.7\textwidth}}
    	\begin{tikzpicture}[scale=.8,every node/.style={minimum size=1cm},on grid]
    	
    	\begin{scope}[
    	yshift=-120,every node/.append style={
    		yslant=0.5,xslant=-1},yslant=0.5,xslant=-1
    	]
    	\fill[white,fill opacity=0.9] (0,0) rectangle (4,4);
    	\draw[step=1cm, black] (0,0) grid (4,4); 
    	\draw[black,very thick] (0,0) rectangle (4,4);
    	\draw[step=1cm, green!90,thick] (3,3) grid (4,4);
    	\fill[green!10!] (3,3) rectangle (4,4);

    	\fill[gray!10!] (0,0) rectangle (1,1);
    	\draw[step=1cm, gray!90,thick] (0,0) grid (1,1);

    	\fill[red!10!] (1,1) rectangle (2,2);
    	\draw[step=1cm, red!90,thick] (1,1) grid (2,2);
    	\fill[red!10!] (2,2) rectangle (3,3);
    	\draw[step=1cm, red!90,thick] (2,2) grid (3,3);
    	
    	\node at (0.75,0.75) {\tiny{$\mathrm{A}$}};
    	\draw [->] (0.5,0.5) -- (0.8,0.5); 
    	\draw [->] (0.5,0.5) -- (0.5,0.8); 
    	\node at (1.75,0.75) {\tiny{{$\mathrm{A}$}}};
    	\draw [->] (1.5,0.5) -- (1.8,0.5); 
    	\draw [->] (1.5,0.5) -- (1.2,0.5); 
    	\node at (2.75,0.75) {\tiny{{$\mathrm{A}$}}};
    	\draw [->] (2.5,0.5) -- (2.8,0.5); 
    	\draw [->] (2.5,0.5) -- (2.2,0.5); 
    	\draw [->] (2.5,0.5) -- (2.5,0.8); 
    	\node at (3.75,0.75) {\tiny{{$\mathrm{A}$}}};
    	\draw [->] (3.5,0.5) -- (3.2,0.5); 
    	\draw [->] (3.5,0.5) -- (3.5,0.8); 
    	\node at (0.75,1.75) {\tiny{{$\mathrm{A}$}}};
    	\draw [->] (0.5,1.5) -- (0.5,1.2); 
    	\draw [->] (0.5,1.5) -- (0.5,1.8); 
    	\node at (1.5,1.5) {$\cdot$};
    	\node at (2.75,1.75) {\tiny{{$\mathrm{A}$}}};
    	\draw [->] (2.5,1.5) -- (2.5,1.2); 
    	\draw [->] (2.5,1.5) -- (2.8,1.5); 
    	\node at (3.75,1.75) {\tiny{{$\mathrm{A}$}}};
    	\draw [->] (3.5,1.5) -- (3.5,1.8); 
    	\draw [->] (3.5,1.5) -- (3.5,1.2); 
    	\draw [->] (3.5,1.5) -- (3.2,1.5); 
    	\node at (0.75,2.75) {\tiny{{$\mathrm{A}$}}};
    	\draw [->] (0.5,2.5) -- (0.5,2.8); 
    	\draw [->] (0.5,2.5) -- (0.5,2.2); 
    	\draw [->] (0.5,2.5) -- (0.8,2.5); 
    	\node at (1.75,2.75) {\tiny{{$\mathrm{A}$}}};
    	\draw [->] (1.5,2.5) -- (1.5,2.8); 
    	\draw [->] (1.5,2.5) -- (1.2,2.5); 
    	\node at (2.5,2.5) {$\cdot$};
    	\node at (3.75,2.75) {\tiny{{$\mathrm{A}$}}};
    	\draw [->] (3.5,2.5) -- (3.5,2.8); 
    	\draw [->] (3.5,2.5) -- (3.5,2.2); 
    	\node at (0.75,3.75) {\tiny{{$\mathrm{A}$}}};
    	\draw [->] (0.5,3.5) -- (0.8,3.5); 
    	\draw [->] (0.5,3.5) -- (0.5,3.2); 
    	\node at (1.75,3.75) {\tiny{{$\mathrm{A}$}}};
    	\draw [->] (1.5,3.5) -- (1.8,3.5); 
    	\draw [->] (1.5,3.5) -- (1.2,3.5); 
    	\draw [->] (1.5,3.5) -- (1.5,3.2); 
    	\node at (2.75,3.75) {\tiny{{$\mathrm{A}$}}};
    	\draw [->] (2.5,3.5) -- (2.8,3.5); 
    	\draw [->] (2.5,3.5) -- (2.2,3.5); 
    	\node at (3.5,3.5) {$\cdot$};
    	
    	\end{scope}
    	
    	\begin{scope}[
    	yshift=0,every node/.append style={
    		yslant=0.5,xslant=-1},yslant=0.5,xslant=-1
    	]
    	\fill[white,fill opacity=.9] (0,0) rectangle (4,4);
    	\draw[black,very thick] (0,0) rectangle (4,4);
    	\draw[step=2cm, black] (0,0) grid (4,4);
    	\node at (1.7,1.75) {\tiny{{$\mathrm{DN}$}}};
    	\draw [->] (1,1) -- (1.3,1); 
    	\draw [->] (1,1) -- (1,1.3); 
    	\node at (3.7,1.75) {\tiny{{$\mathrm{DN}$}}};
    	\draw [->] (3,1) -- (3,1.3); 
    	\draw [->] (3,1) -- (2.7,1); 
    	\node at (1.7,3.75) {\tiny{{$\mathrm{DN}$}}};
    	\draw [->] (1,3) -- (1.3,3); 
    	\draw [->] (1,3) -- (1,2.7); 
    	\node at (3.7,3.75) {\tiny{{$\mathrm{DN}$}}};
    	\end{scope}

    	
    	\node at (-6.2,2) {$\mathsf{Level~1}$};
    	\node at (-6.2,-3) {$\mathsf{Level~2}$};
    	
    	\end{tikzpicture}
	\end{adjustbox}
	\caption{Visualization of the available actions at each state.  Actions ``$\mathrm{A}$" and ``$\mathrm{DN}$" represent abstract up and abstract down, respectively.  Arrows indicate movements within a given level.}
	\label{fig:availact}
\end{figure}
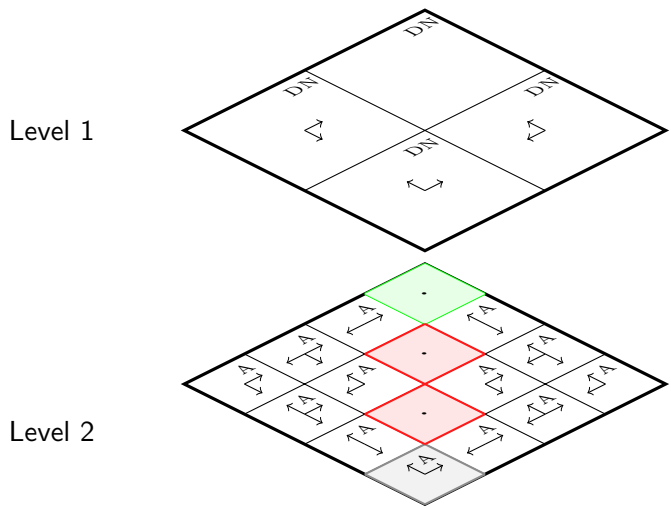

We demonstrate the previous algorithms by applying them to a path-planning problem in a grid world.
Consider an agent  navigating over a 4 $\times$ 4 square (the world environment) that is full of obstacles, as shown in
Fig.~\ref{fig:gridex11}.
The goal is to safely navigate from the start state (cell 1) to the goal state (cell 16) while avoiding the obstacles.
At each time step $t=0,1,\ldots$ the agent can choose its actions from the set $U(x_t) \subseteq \{\textrm{UP, DN, L, R, A-UP, A-DN}\}$.
The actions A-UP and A-DN do not correspond to physical movement, but rather encode changes in the ``perceptual representation'' of the environment that are motivated by the sensing and/or computational resources of the agent.
Note that not all actions are available at every state, as shown in Fig.~\ref{fig:availact}.

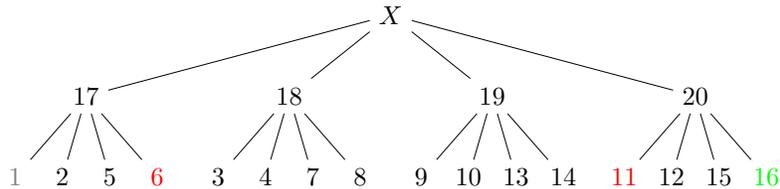
\begin{figure}
	\centering
	\begin{adjustbox}{max size={0.75\textwidth}}
	\begin{tikzpicture}[level distance=1.2cm,
	level 1/.style={sibling distance=3cm},
	level 2/.style={sibling distance=0.7cm}]
	\node {$X$}
	child {node {17}
		child {node {\color{gray}1\color{black}}}
		child {node {2}}
		child {node {5}}
		child {node {\color{red}6\color{black}}}
	}
	child {node {18}
		child {node {3}}
		child {node {4}}
		child {node {7}}
		child {node {8}}
	}
	child {node {19}
		child {node {9}}
		child {node {10}}
		child {node {13}}
		child {node {14}}
	}
	child {node {20}
		child {node {\color{red}11\color{black}}}
		child {node {12}}
		child {node {15}}
		child {node {\color{green}16\color{black}}}
	};
	\end{tikzpicture}
	\end{adjustbox}
	\caption{Hierarchical description of states with coloring consistent with description in Fig. \ref{fig:gridex11}.}
	\label{fig:quadtree}
\end{figure}

The space the agent operates in (and reasons about) is an augmented grid world with two abstraction levels, as shown in Fig.~\ref{fig:gridex11}.
Specifically, the world states 1,2,5,6 are all abstracted by the additional new (perceptual) state 17, states 3,4,7,8 are all abstracted by the new state 18 and so on, as shown in Figs.~\ref{fig:gridex11} and \ref{fig:quadtree}.
Similarly, perceptual states 17,18,19,20 can all be abstracted, in turn, by the new (root) state 21 (X).
This situation is similar to a standard quad-tree decomposition for this grid world scenario, although one can envision different methods to generate the corresponding abstractions.
Here we do not allow the agent to abstract to the root state, and hence the final search space includes 20 states.

For each state $x \in \{ 1,2,\ldots, 20\}$ the set of available actions $U(x) \subseteq \{ \textrm{UP}, \textrm{DN}, \textrm{L}, \textrm{R}, \\ \textrm{A-UP}, \textrm{A-DN}\}$
includes only actions that will not \emph{intentionally} move the agent into an obstacle (see Fig. \ref{fig:availact}).
Since the transitions are generally probabilistic, there is a possibility that the agent may end up in \emph{any} neighboring state.
Thus, actions have a success probability of 80\%, with the remaining 20\% distributed among the neighboring states.
For example, selecting the action UP in state 1 will with 80\% probability move the agent to state 5, while the remaining 20\% is distributed over states 2 and 6 using this action.
The exception to this rule are the special actions A-UP and A-DN.
These special actions are deterministic and can be viewed as the agent's ability to toggle a sensor suite consisting of two variable resolution sensors on and off.
The agent must pay an increased price for navigating the grid world at the finest resolution level, due to the demand to process an increased amount of information.
The actions A-UP and A-DN can thus also be viewed as actions that are taken when the agent has limited computational resources, in which case it will favor higher level (coarser) abstractions since these correspond to search spaces of lower cardinality and hence are presumably easier to navigate in.

In terms of transitions for special actions A-UP and A-DN, we assume that if an agent abstracts up, it will find itself in the corresponding aggregate state at the next epoch.
Conversely, if an agent decides to abstract down, it will find itself in one of the underlying states with equal probability at the next time step.
For example, executing A-UP from state 1 will deterministically move the agent to state 17, while executing A-DN from state 17 will move the agent to
either state 1, 2, 5, or 6 with equal probability.

The rewards function is constructed as follows.  The agent pays a cost of -0.55 for every action corresponding to UP, DN, L, R within level 2, and -0.15 for these actions in level 1.
To abstract up, the agent must pay a cost of approximately -2, and -1 to abstract down.
It receives +2 for reaching the goal and -10 for navigating into an obstacle.
The discount factor is 0.95.
It should be noted that goal and obstacles are \emph{absorbing} states, hence navigation into obstacles is catastrophic.
Also, since goal state 16 is accessible from state 20, we assume that the reward the agent receives for this transition is $R_{\mathrm{abstract DN}} + R_{\mathrm{goal}}$ to correspond to both the price of switching to the higher resolution sensor and finding itself at the goal state.

\begin{figure}[htb]
	\centering 
		\subfigure[]{\includegraphics[scale = 0.13]{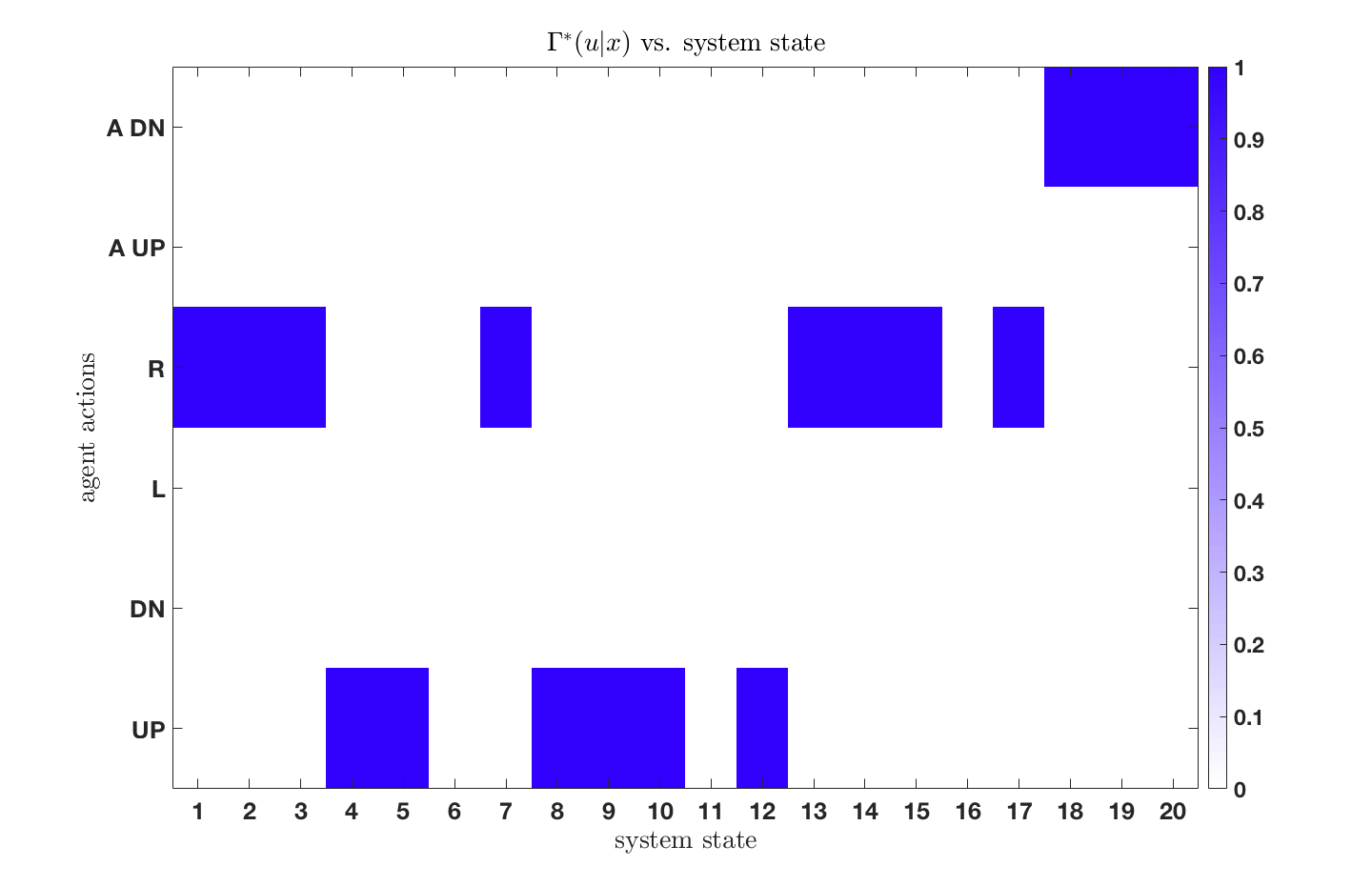}}
		\hspace*{-1.0mm}
		\subfigure[]{\includegraphics[scale = 0.13]{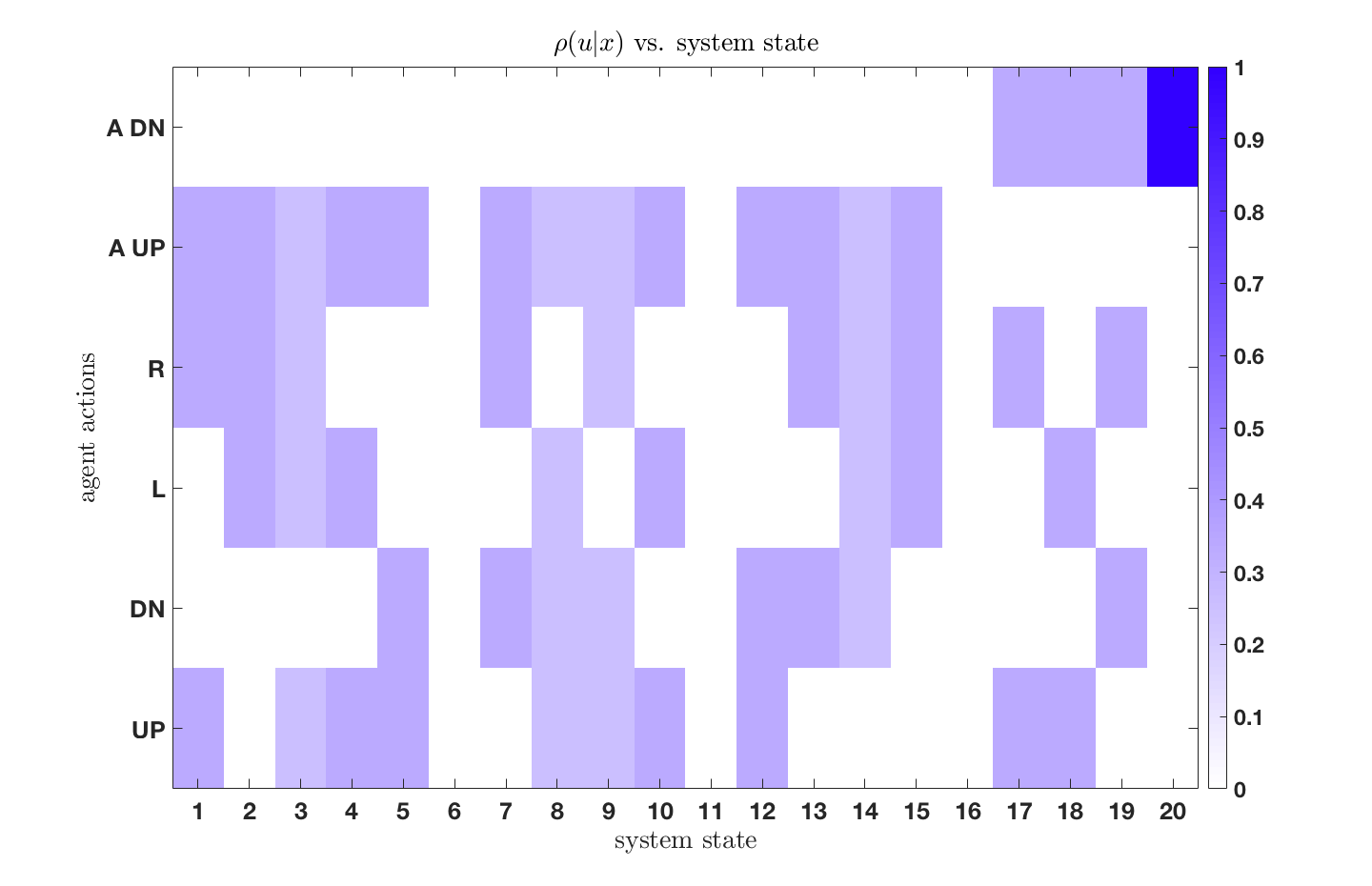}}
	\caption{Deterministic policy, $\Gamma^{*}(u|x)$ (left) and a priori action distribution, $\rho(u|x)$ (right).}
	\label{fig:gamma_and_rho}	
\end{figure}

In the discussion that follows all quantities that refer to $x_0$ are the values of the variable at state 1.
For example, $V^{\pi}(x_0)$ is the expected value of starting from state 1 and following the given policy $\pi(u|x)$.
As a measure to compare the performance of the other algorithms and to better understand how the resource parameter alters the agent's behavior, we first present results for the deterministic case ($\beta \to \infty$).
The deterministic policy obtained from solving the standard MDP problem is shown in Fig.~\ref{fig:gamma_and_rho}a and has an associated value of $V^{\pi}(x_0) = -4.22$.
It is important to note that the policy depicted in Fig.~\ref{fig:gamma_and_rho}a by $\Gamma^{*}$ represents a {rational} agent.
That is,  the policy $\Gamma^{*}$ would be followed by an agent who has sufficient processing capabilities to find the optimal actions and does so on the finest grid level, electing to not abstract to a coarser resolution for path planning.
It will be seen that, as the agent's computational abilities are reduced, it begins to favor planning at the coarser grid level while its resulting policy becomes increasingly suboptimal with respect to $\Gamma^{*}$.

\begin{figure}[htb]
	\centering 
	\subfigure[]{\includegraphics[scale = 0.13]{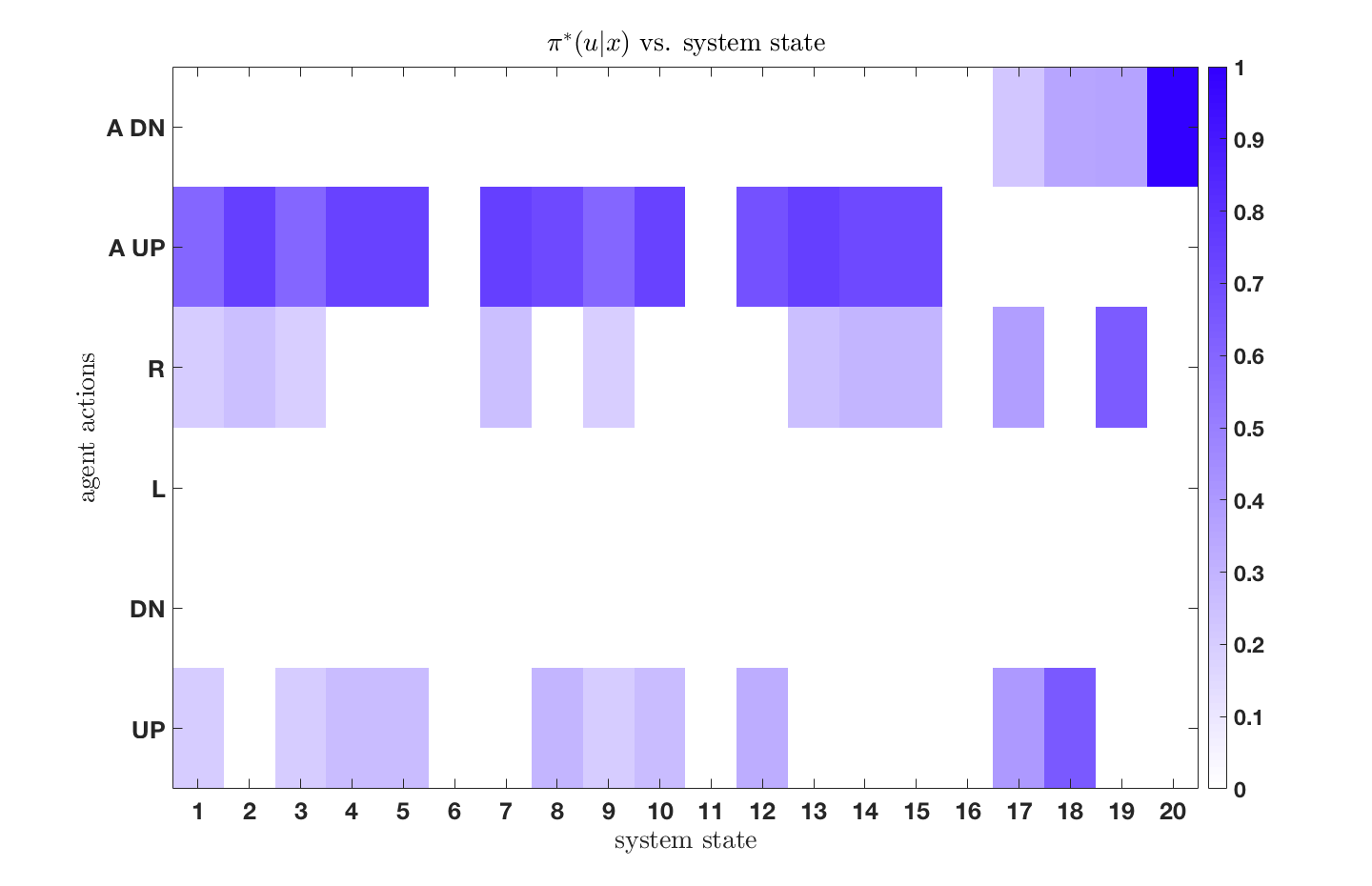}}
	\hspace*{-1.0mm} %
	\subfigure[]{\includegraphics[scale = 0.13]{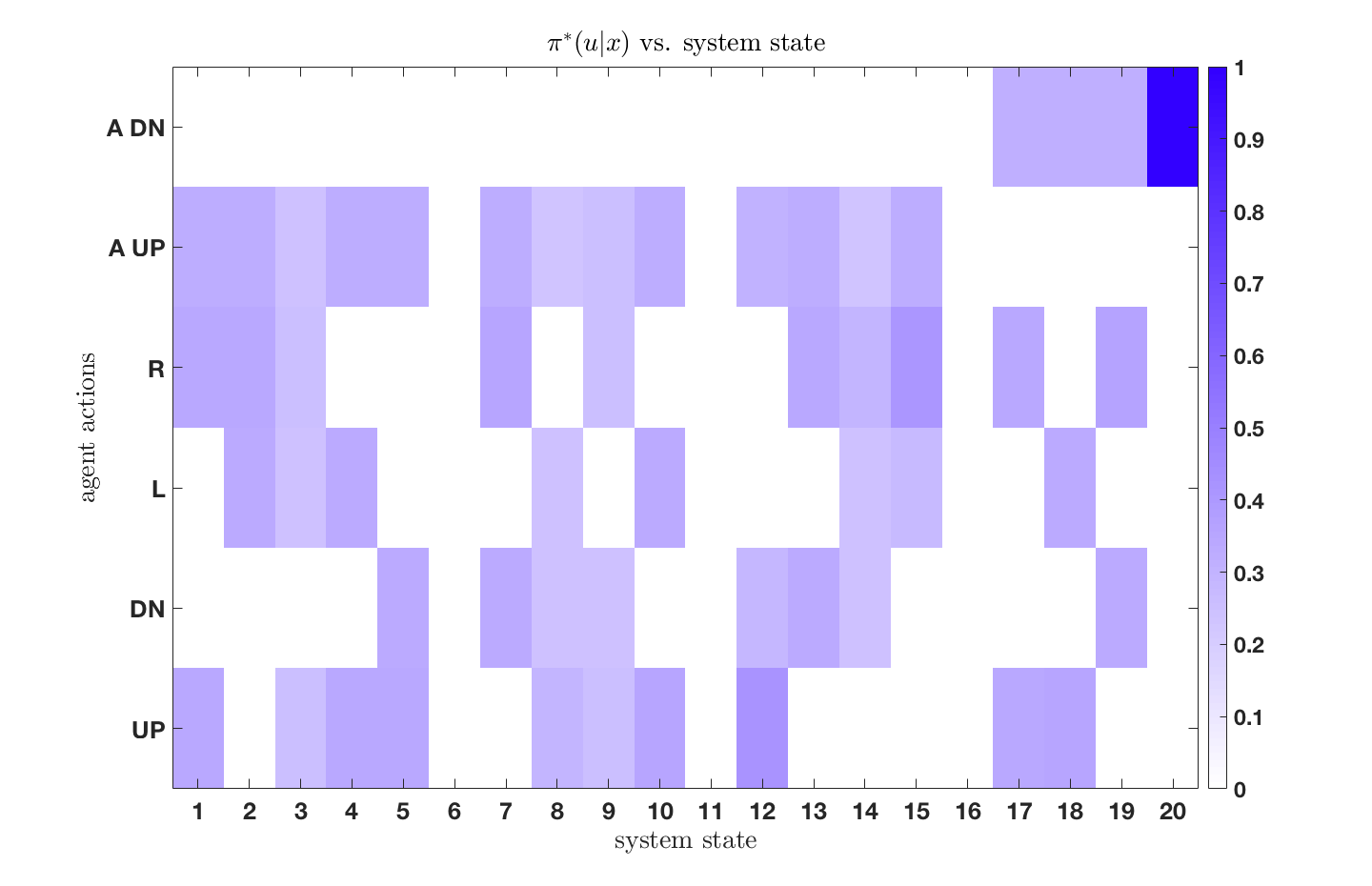}}
	\caption{$\pi^{*}(u|x)$ from the BA-IL algorithm (left) and $\pi^{*}(u|x)$ from the IL algorithm (right) for $\beta$ = 0.05.}
	\label{fig:expl4}	
\end{figure}

Shown in Fig.~\ref{fig:gamma_and_rho}b is the prior action distribution, $\rho(u|x)$.
The prior distribution, $\rho$, can be viewed as the policy the agent would follow in case it had no resources to find another, possibly better, policy \cite{Rubin12}.
Thus, if the agent has insufficient resources to deliberate (limit as $\beta{} \to 0$) , $\rho$ becomes the posterior policy $\pi$, as seen when comparing Figs.~\ref{fig:gamma_and_rho}b and \ref{fig:expl4}b.

\begin{figure}[htb]
	\centering 
	\subfigure[]{\includegraphics[scale = 0.13]{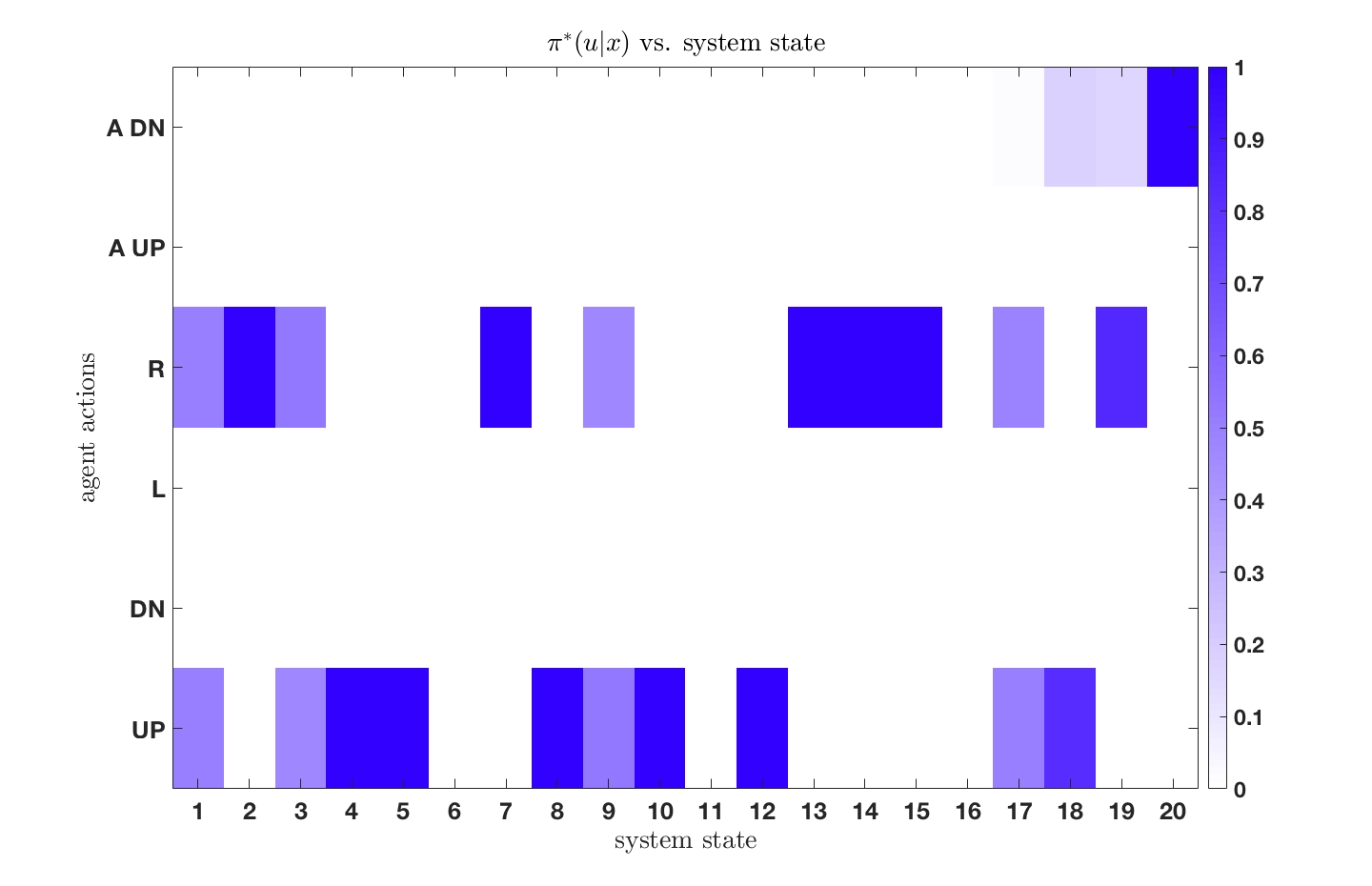}}
	\hspace*{-1mm}
	\subfigure[]{\includegraphics[scale = 0.13]{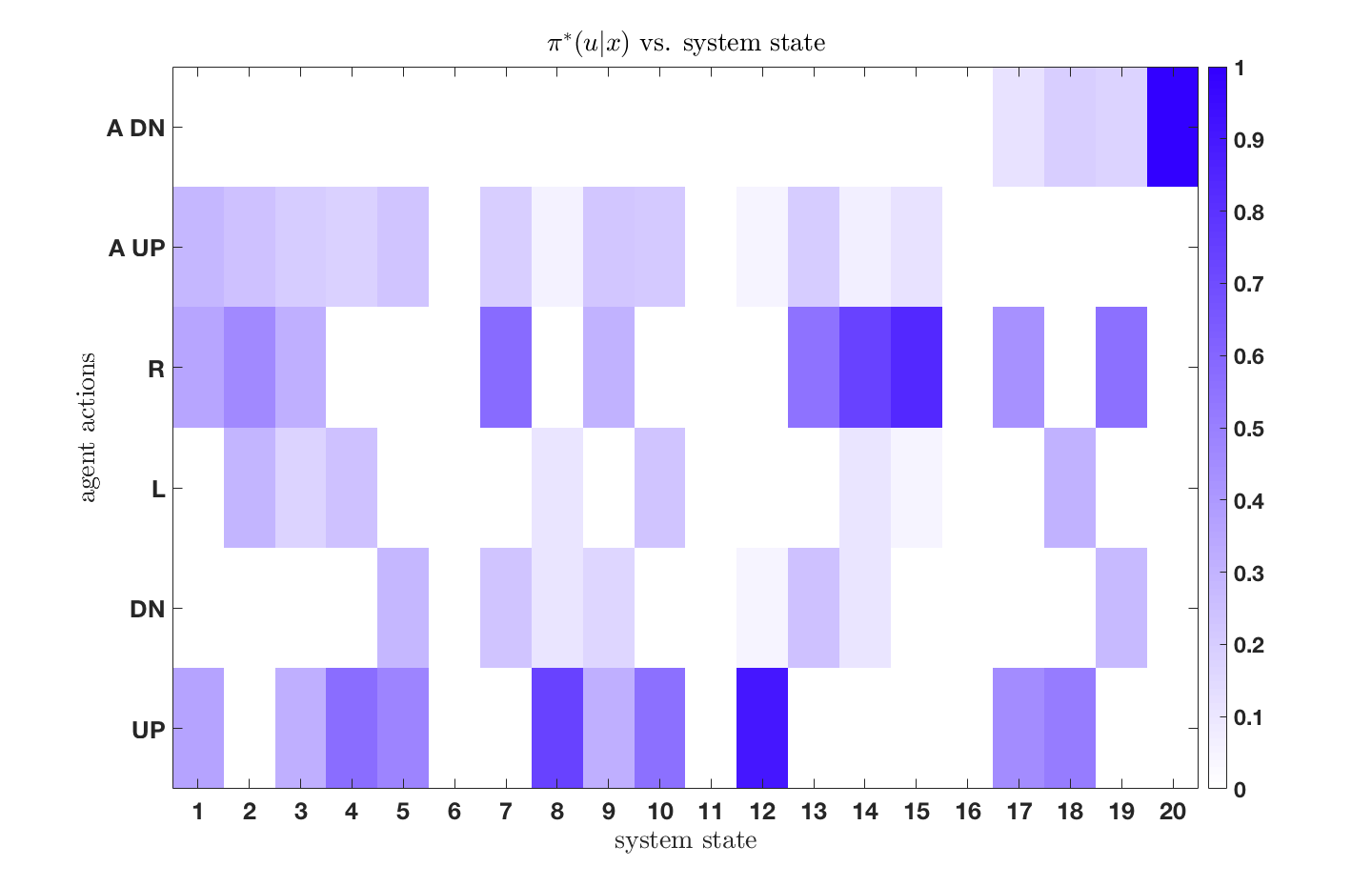}}
	\caption{$\pi^{*}(u|x)$ from the BA-IL algorithm (left) and $\pi^{*}(u|x)$ from the IL algorithm (right) for $\beta$ = 0.5.}
	\label{fig:expl2}	
\end{figure}

Figures~\ref{fig:gamma_and_rho}-\ref{fig:expl4} display the policies obtained for various values of $\beta$ for both the
information-limited (IL) and the Blahut-Arimoto information limited (BA-IL) algorithms.
More precisely, the figures show the distribution over actions for a given state and value of $\beta$ (and the distribution over states, $p(x)$, in the case of the BA algorithm).
In each of these figures, the y-axis corresponds to the action space whereas the x-axis displays the possible system states, with state numbering consistent with Fig.~\ref{fig:gridex11}.
Furthermore, the shading of a given state-action pair represents the probability that the action will be selected when the system finds itself in that state.
All plots are configured so that the darker shading represents a higher probability of an action being selected when following the policy, with no shading representing zero probability.
Column $n$ of these plots can therefore be considered a probability distribution over actions for the $n^{\text{th}}$ state, where $n \in \{ 1,2,\ldots, 20\}$.

Table~\ref{ta:v_pi_ex1} shows the corresponding values of following policies obtained from the algorithms for a variety of $\beta{}$.
\begin{table}[t]
		\caption{Value of policy $\pi^{*}$ for various values of $\beta$.}
		\label{ta:v_pi_ex1}
		\centering
		\begin{tabular}{c|c|c}
			$V^{\pi}(x_0)$ IL-BA & $V^{\pi}(x_0)$ IL & \textbf{$\beta$} \\ \hline
			-10.5 & -11 & 0.05 \\ \hline
			-4.33 & -7.39 & 0.5 \\ \hline
			-4.27 & -4.35 & 5 \\ \hline
			-4.22 & -4.22 & 100 \\ \hline
		\end{tabular}
\end{table}
Note that as $\beta \to \infty$ the value of the policy $\pi^{*}$ approaches that of $\Gamma^{*}$, although does so non-linearly \cite{Rubin12}.
This is also observed in Figs.~\ref{fig:expl2}-\ref{fig:expl4} as the resulting policies ($\pi^*$) approach $\Gamma^{*}$ as $\beta$ is increased.
We also note that the policy becomes less deterministic as $\beta \to 0$.
Furthermore, as $\beta$  is reduced, abstractions emerge from the solution of the BA algorithm.
This can be seen in Fig.~\ref{fig:expl4}a since, when compared to the non-BA policy in Fig.~\ref{fig:expl4}b, we see that the agent elects to select the same action in many states.
That is, the resulting optimal policy $\pi^{*}$ from the BA algorithm is relatively \emph{generic} with respect to the state and thus the state and action spaces have low mutual information.
Because of the additional step in optimizing the prior distribution in the BA algorithm, a resource restricted agent following a BA optimal policy achieves higher value for a given $\beta$ when compared to the same agent following an optimal policy provided by the information-limited algorithm, as seen in Table \ref{ta:v_pi_ex1}.
Finally, it should be noted from Fig.~\ref{fig:expl4} that as the information parameter is reduced, the agent begins to path plan at the coarser grid level, indicative that less computationally complex solutions are sought at the cost of complete optimality.

\section{CONCLUSION}  \label{sec:conclusion}

We have presented an overview of an information-theoretic approach to account for computational resources in sequential decision making problems modeled by MDPs.
It is shown how the limited information processing costs give rise to state hierarchical abstractions, which can then be used to solve the decision problem at the best state representation ``granularity'' level.
We have demonstrated how the approach can be applied to a typical path-planning problem.
It is shown that the resulting optimal policy changes significantly as a function of the processing constraints.
In the process, we proposed a simple extension to the classical Blahut-Arimoto algorithm from rate distortion theory that is useful in many engineering applications.


\bibliographystyle{IEEEtran}
\bibliography{msbib}
\end{document}